\PassOptionsToPackage{colorlinks=true,linkcolor=linkpink,citecolor=linkpink,urlcolor=linkpink}{hyperref}

\documentclass[runningheads]{llncs}
\usepackage[T1]{fontenc}
\usepackage{graphicx}
\usepackage{subcaption}
 \usepackage{booktabs}
\usepackage{float}
\usepackage{color}
\usepackage{tcolorbox}
\usepackage{amsfonts,amsmath,amssymb}
\usepackage{pgfplots}
\usepackage{cite}
\usepackage{breqn}
\usepackage{booktabs}
\usepackage{soul}
\usepackage{paralist}
\usepackage{enumerate}
\usepackage{wrapfig}

\usepackage{xargs}                      
\pgfplotsset{compat=1.3}

\usepackage{pifont}%

\emergencystretch=\hsize
\tikzstyle{fancylabel} = [text=yellow, inner sep=0pt, minimum size=15pt, align=left]
\tikzstyle{mylabel} = [text=white, ultra thick, inner sep=1pt, minimum size=15pt, yshift=-4pt, xshift=6pt]

\makeatletter
\DeclareRobustCommand\onedot{\futurelet\@let@token\@onedot}
\def\@onedot{\ifx\@let@token.\else.\null\fi\xspace}

\makeatother

\newcommand{\mytodo}[1]{\todo{TODO}}

\label{packages}
\usepackage{environ}

\usepackage{color}
\usepackage{enumitem}
\usepackage{graphicx}
\usepackage{grffile}
\usepackage{marvosym}
\usepackage{complexity}
\usepackage{algorithm}

\usepackage{tcolorbox}
\usepackage{url}
\usepackage{graphicx}
\usepackage{color}
\usepackage{algorithm,algorithmic}

\usepackage{dsfont} \usepackage{pifont} \usepackage{bbm}
\usepackage{comment}
\usepackage{rotating}

\definecolor{babyblue}{rgb}{0.54, 0.81, 0.94}
\definecolor{armygreen}{rgb}{0.29, 0.33, 0.13}
\definecolor{brightlavender}{rgb}{0.75, 0.58, 0.89}
\definecolor{aqua}{rgb}{0.0, 1.0, 1.0}
\definecolor{caribbeangreen}{rgb}{0.0, 0.8, 0.6}
\definecolor{reddish}{rgb}{0.82, 0.1, 0.26}
\definecolor{emerald}{rgb}{0.31, 0.78, 0.47}
\definecolor{jasper}{rgb}{0.84, 0.23, 0.24}
\definecolor{red}{rgb}{1.0, 0.0, 0.0}
\definecolor{green}{rgb}{0.0, 1.0, 0.0}
\definecolor{blue}{rgb}{0.0, 0.0, 1.0}
\definecolor{darkgreen}{rgb}{0.1, 0.7, 0.1}
\definecolor{darkblue}{rgb}{0.1, 0.1, 0.7}
\definecolor{red}{rgb}{0.7, 0.1, 0.1}
\usepackage{graphicx,verbatim}
\usepackage{orcidlink}

\usepackage{xcolor}
\definecolor{linkpinkix}{HTML}{EA335A} 

\definecolor{linkpink}{HTML}{EA335A}

\usepackage{tikz}

\usetikzlibrary{arrows}

\graphicspath{{./}}

\usepackage[align=center,shadow=true,shadowsize=5pt,nobreak=true,framemethod=tikz,style=0,skipabove=2pt,skipbelow=1pt,innertopmargin=-3pt,innerbottommargin=3pt,innerleftmargin=5pt,innerrightmargin=5pt,leftmargin=-2pt,rightmargin=-2pt]{mdframed}

\usepackage[colorlinks=true,linkcolor=linkpink,citecolor=linkpink]{hyperref}

\usetikzlibrary{shadows}
\usepackage{color}

\usepackage{tikz,xcolor}
 
\definecolor{lime}{HTML}{A6CE39}
\DeclareRobustCommand{\orcidicon}{
	\begin{tikzpicture}
	\draw[lime, fill=lime] (0,0) 
	circle [radius=0.16] 
	node[white] {{\fontfamily{qag}\selectfont \tiny ID}};
	\draw[white, fill=white] (-0.0625,0.095) 
	circle [radius=0.007];
	\end{tikzpicture}
	\hspace{-2mm}
}
 
\foreach \x in {A, ..., Z}{\expandafter\xdef\csname orcid\x\endcsname{\noexpand\href{https://orcid.org/\csname orcidauthor\x\endcsname}
			{\noexpand\orcidicon}}
}

\usepackage{tikz}
\usepackage{xparse}   

\DeclareRobustCommand{\authorpic}[2][5mm]{%
  \tikz[baseline={([yshift=-.25ex]current bounding box.center)}]{%
    \clip (0,0) circle (#1);
    \pgfmathsetlengthmacro{\picside}{sqrt(2)*#1}%
    \node at (0,0) {\includegraphics[width=\picside,height=\picside,keepaspectratio]{#2}};
    \draw[line width=0.4pt, color=white] (0,0) circle (#1);
  }%
}

\NewDocumentCommand{\AuthorWithPic}{O{5.5mm} O{0.20em} m m}{%
  \texorpdfstring{\authorpic[#1]{#4}\kern #2}{}%
  #3%
}

\newcommand{\shadedlink}[2]{%
  \tikz[baseline=(n.base)]\node[
    fill=linkpink,
    fill opacity=0.5,
    text opacity=1,
    rounded corners=.3ex,
    inner xsep=.35em,
    inner ysep=.15em
  ] (n) {\href{#1}{\textcolor{blue!70!black}{#2}}};%
}

\begin{document}
\title{GNN-based Unified Deep Learning}
%
%

\author{%
  \AuthorWithPic[6mm][0.18em]{Furkan Pala\orcidA{}}{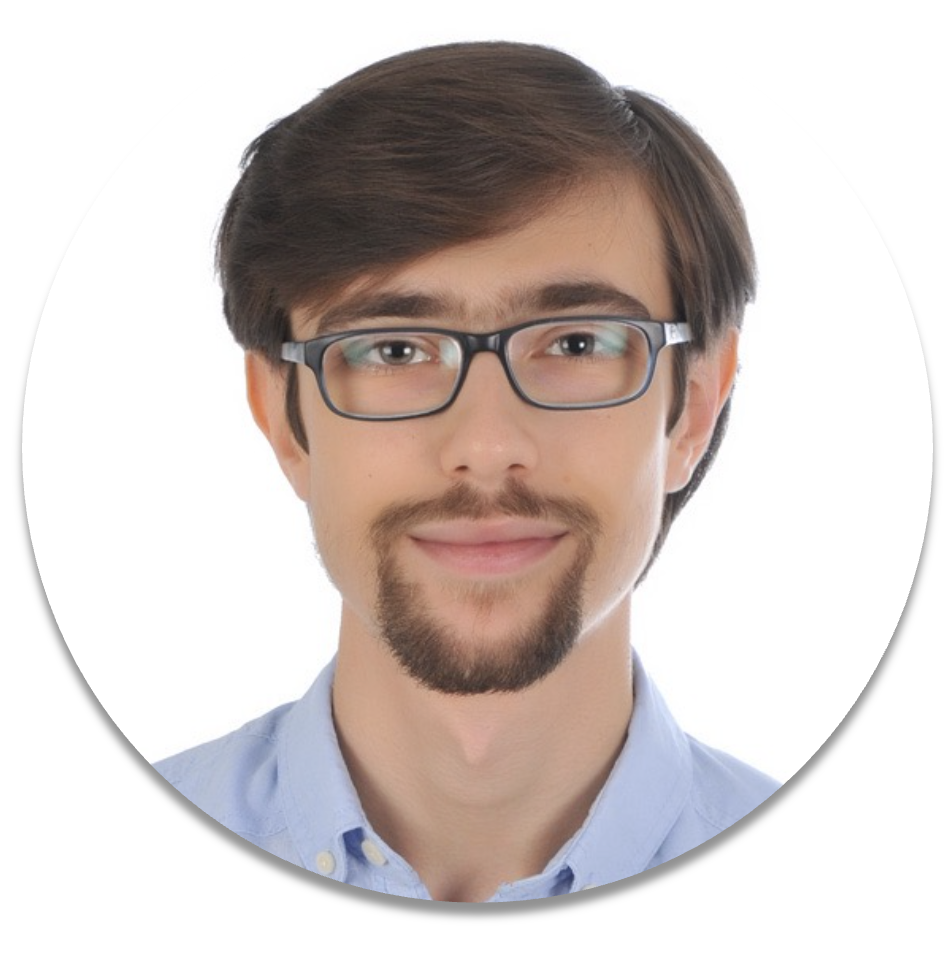} \and
  \AuthorWithPic[6mm][0.18em]{Islem Rekik\orcidB{}}{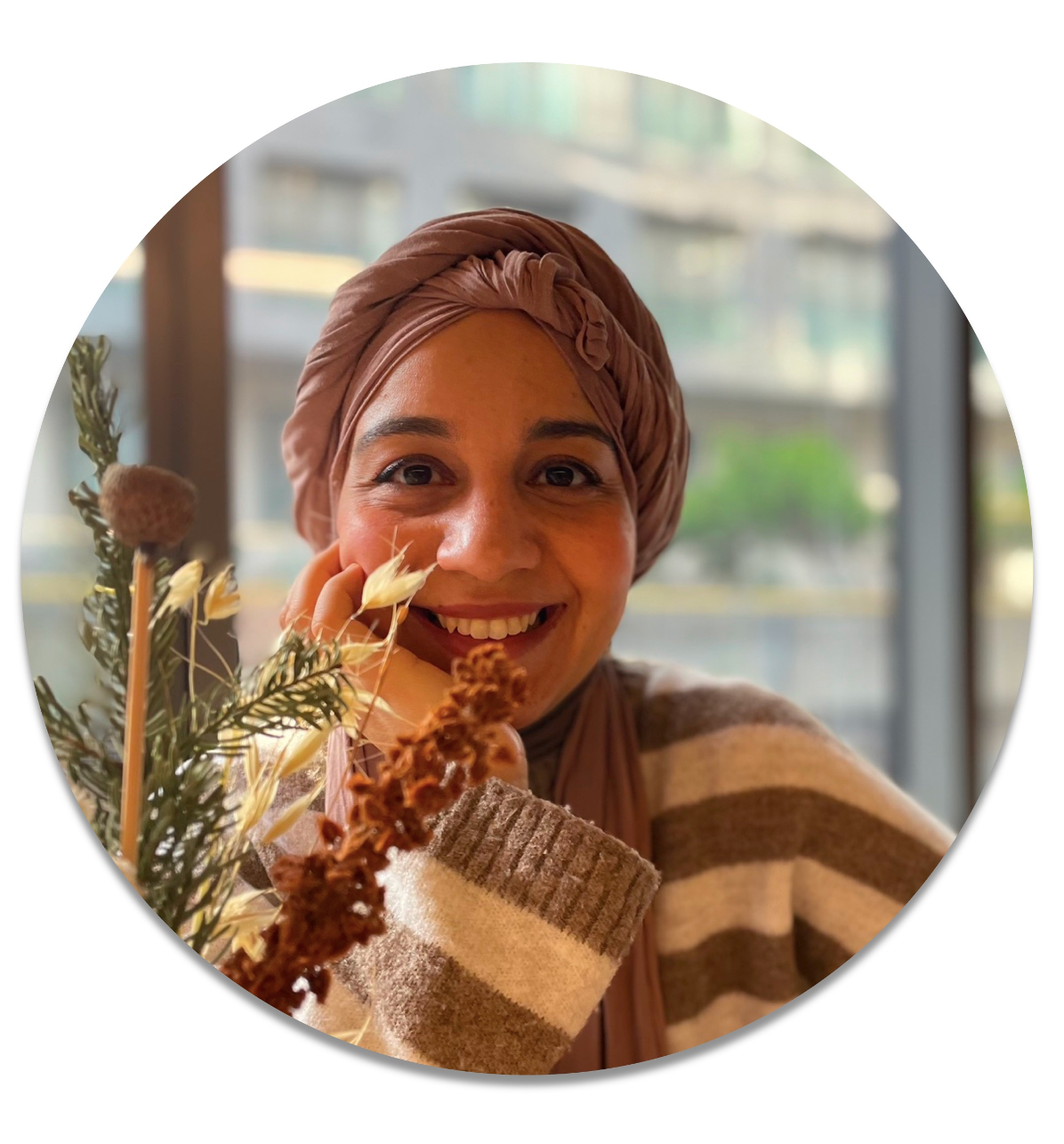}\thanks{Corresponding author: \email{f.pala23@imperial.ac.uk,i.rekik@imperial.ac.uk}, \url{http://basira-lab.com}, GitHub: \url{https://github.com/basiralab/uGNN}}%
}

\institute{BASIRA Lab, Imperial-X (I-X) and Department of Computing, Imperial College London, London, United Kingdom}
\authorrunning{Pala et al.}

\maketitle    

\begin{abstract}
Deep learning models often struggle to maintain robust generalizability in medical imaging, particularly under \emph{domain-fracture} scenarios where distributional shifts arise due to varying imaging techniques, acquisition protocols, patient populations, demographics, and equipment. In practice, each hospital may need to develop and train distinct models---differing in functionality (i.e., learning task) and morphology including width and depth---to handle their local data distributions. For example, while one hospital may utilize Euclidean architectures such as MLPs and CNNs to process structured tabular data or regular grid-like image data, another hospital may need to deploy non-Euclidean architectures such as graph neural networks (GNNs) to process inherently irregular data like brain connectomes or other graph-structured biomedical information. However, how to train such heterogeneous models coherently across different datasets, in a manner that enhances the generalizability of each model, remains an open and challenging problem. In this paper, we address this issue by introducing a new learning paradigm, namely \textbf{unified learning}. To address the topological differences between these heterogeneous architectures, we first encode each model into a graph representation, enabling us to unify these diverse models within a shared graph learning space. Once represented in this space, a GNN guides the optimization of the unified models. By decoupling the parameters of individual deep learning models and controlling them through the unified GNN (\texttt{uGNN}), our approach enables parameter-sharing and knowledge-transfer across \emph{varying} architectures (MLPs, CNNs and GNNs) and distributions, ultimately improving its generalizability. We evaluate our framework on MorphoMNIST and two MedMNIST benchmarks---PneumoniaMNIST and BreastMNIST---and find that our unified learning improves the performance of individual models when trained on unique distributions and tested on mixed ones, thereby demonstrating generalizability to unseen data with strong distributional shifts. Our source code including benchmarks and evaluation datasets is available at \url{https://github.com/basiralab/uGNN}. \footnote{This paper has been selected for an \textbf{Oral Presentation} at the PRIME MICCAI 2025 workshop. \shadedlink{https://youtu.be/PnlrI-6ykTY}{[uGNN YouTube Video]}.}

\end{abstract}

\section{Introduction}

Deep learning has led to significant advances in medical image analysis, improving accuracy and robustness in diverse tasks such as lesion detection, organ segmentation, and disease classification~\cite{litjens2017survey,shen2017deep,yan2018deeplesion}. Despite such successes, achieving robust generalization across varying clinical conditions remains challenging, particularly in \textit{domain-fracture} scenarios~\cite{yoon2024domain,nguyen2024domain}. Such scenarios arise when data are drawn from heterogeneous sources---e.g. different scanners, patient populations, or imaging protocols---leading to distributional shifts between the train and test datasets that degrade performance. Conventional strategies, such as domain adaptation or federated learning~\cite{li2020review,kairouz2021advances}, often assume architectural homogeneity or alignments between models, thus limiting their capacity to handle fundamentally heterogeneous deep learning architectures.

In practice, healthcare institutions develop machine learning models with differing architectures and parametrizations, tailored to their local patient populations or imaging modalities~\cite{wen2023source}. These heterogeneous architectures can vary from simpler Multi-Layer Perceptrons (MLPs) to more complex Convolutional Neural Networks (CNNs) and Graph Neural Networks (GNNs), each trained under different conditions. Integrating knowledge across these models is challenging, as their internal representations, layer connectivity, and parameters can differ significantly. Existing learning paradigms are not inherently designed to reconcile such differences, leaving a critical gap in translating learned knowledge across diverse models and data distributions.

To address these limitations, we introduce the \emph{unified learning} framework, which unifies the learning process across highly heterogeneous deep learning architectures within a common representational space. Unlike traditional approaches such as transfer learning, knowledge distillation, and ensemble learning, which focus on aligning parameters, distilling knowledge, or combining model outputs, unified learning represents each model—regardless of architecture—as a graph. This graph captures the model's layers, connections, and parameters as nodes and edges. By embedding these heterogeneous model-graphs into a shared space, a single graph neural network (GNN) drives their learning, optimizing the GNN instead of directly updating individual model parameters.

\begin{mdframed}[frametitle={\colorbox{white}{\space Hypothesis:\space}},%
frametitleaboveskip=-\ht\strutbox,%
frametitlealignment=\center]
We hypothesize that \textbf{unified learning} solves the generalizability of heterogeneous architectures under domain-fracture scenarios, enabling models to adapt robustly to distributional shifts without enforcing architectural uniformity.
\end{mdframed}

The key challenge then becomes how to integrate heterogeneous model representations into a common graph space using a GNN:

\begin{mdframed}[frametitle={\colorbox{white}{\space Key Challenge:\space}},%
frametitleaboveskip=-\ht\strutbox,%
frametitlealignment=\center]
How can we unify heterogeneous deep learning architectures within a common graph space, and effectively bridge the gap between Euclidean models (MLPs, CNNs) and non-Euclidean models (GNNs), when training across diverse datasets using a GNN?
\end{mdframed}

By formulating heterogeneous models as graphs and training a GNN to orchestrate their parameter updates, we establish a paradigm that overcomes the limitations imposed by architectural homogeneity. This approach harnesses the topological flexibility of graph representations to accommodate diverse model architectures, enabling both efficient knowledge sharing and enhanced adaptation to new domains. In summary, our key contributions are as follows: \textbf{(1)} We propose a \textit{unified learning} paradigm that represents heterogeneous deep learning architectures as graphs, enabling a GNN to simulate forward passes and parameter updates for varied models in a common learning space. \textbf{(2)} Our unified learning framework allows training Euclidean and non-Euclidean models coherently. \textbf{(3)} We address the inherent complexity of domain-fracture scenarios by integrating distinct models and datasets into a single, architecture-agnostic training process, thereby facilitating robust cross-domain generalization. \textbf{(4)} We provide empirical evidence on medical imaging benchmarks, demonstrating that unified learning improves performance, stability, and adaptability of heterogeneous models facing distributional shifts.

\section{Related Work}

Deep learning has revolutionized medical image analysis, enabling breakthroughs in disease detection, segmentation, and classification tasks~\cite{litjens2017survey,shen2017deep}. However, generalizing across domain-fracture medical imaging datasets, characterized by heterogeneity in imaging protocols, scanner settings, and patient demographics, remains a critical challenge~\cite{kamnitsas2017unsupervised}. Traditional learning paradigms such as transfer learning, knowledge distillation, and federated learning have been extensively explored to address domain fracture but are constrained by their inability to accommodate heterogeneous deep learning architectures~\cite{hinton2015distilling}.

\textbf{Domain Fracture in Medical Imaging.}
Domain shifts in medical imaging arise due to differences in acquisition settings, modality characteristics, and institutional practices~\cite{yoon2024domain}. Federated learning (FL) has emerged as a prominent strategy for collaboratively training models across decentralized datasets while maintaining data privacy~\cite{yang2019federated}. In the context of medical imaging, FL has shown considerable promise for addressing domain shifts and enhancing model generalization across decentralized datasets, making it a rapidly growing area of research with numerous advancements~\cite{sohan2023systematic,rehman2023federated}. However, most FL methods assume homogeneous architectures across participating institutions~\cite{kairouz2021advances}. Similarly, domain adaptation techniques, such as adversarial learning-based approaches, aim to align feature spaces across domains~\cite{ganin2015unsupervised,zhou2020learning}. While effective in some cases, these approaches often require data centralization or strong assumptions about the underlying data distributions, making them impractical for highly fractured medical imaging data.

\textbf{Learning Paradigms for Heterogeneous Architectures.}
The challenge of synergizing and integrating the training of heterogeneous architectures has received limited attention. For instance, knowledge distillation~\cite{hinton2015distilling} enables the transfer of knowledge from a complex teacher model to a simpler student model but relies on aligning output distributions rather than internal representations. Collaborative learning approaches~\cite{singh2020model} have explored model aggregation but remain limited to networks with similar architectures. In contrast, our proposed model \emph{unifies} heterogeneous architectures into a common graph-based learning space, enabling shared learning dynamics across disparate networks as well as datasets.

\textbf{Graph Neural Networks in Medical Imaging.}
Graph-based methods have gained traction in medical imaging for their ability to model complex spatial and relational structures~\cite{bronstein2017geometric}. Cell-graph representations, which capture cellular interactions and tissue structures as graphs, have shown promise in histopathology and disease characterization tasks~\cite{wang2023review}. GNNs extend this framework by learning representations over graphs, enabling tasks such as tumor classification~\cite{ravinder2023enhanced} and disease progression modeling~\cite{zou2022modeling}. However, to the best of our knowledge, GNNs have never been leveraged to harmonize the training and learning of heterogeneous deep neural networks (DNNs).

\section{Methodology}

This section presents our \texttt{uGNN} framework, which unifies the learning of heterogeneous deep learning architectures within a shared graph-based representation space. In our framework, we can train MLPs, CNNs and GNNs coherently, making unified learning a powerful framework to train Euclidean and non-Euclidean models together. 

\textbf{Motivation.} Our motivation for using graphs and GNNs stems from the structural heterogeneity of deep learning architectures. Each model type—such as MLPs, CNNs, or GNNs—differs in terms of its layers, parameterization, and neuron connectivity. To unify these architectures under a common representation, we convert each of them into a graph structure. These resulting \emph{model-graphs} abstract away architectural differences. Once in graph form, we employ a GNN to process and learn over these model-graphs. The GNN serves as a harmonizing function that operates over all trainable parameters across heterogeneous architectures within a shared graph-based representation space. This allows the models to co-train, exchange knowledge, and generalize more effectively—particularly under domain shifts or non-IID training conditions.

\begin{figure}[ht]
    \centering
    \includegraphics[width=1\linewidth]{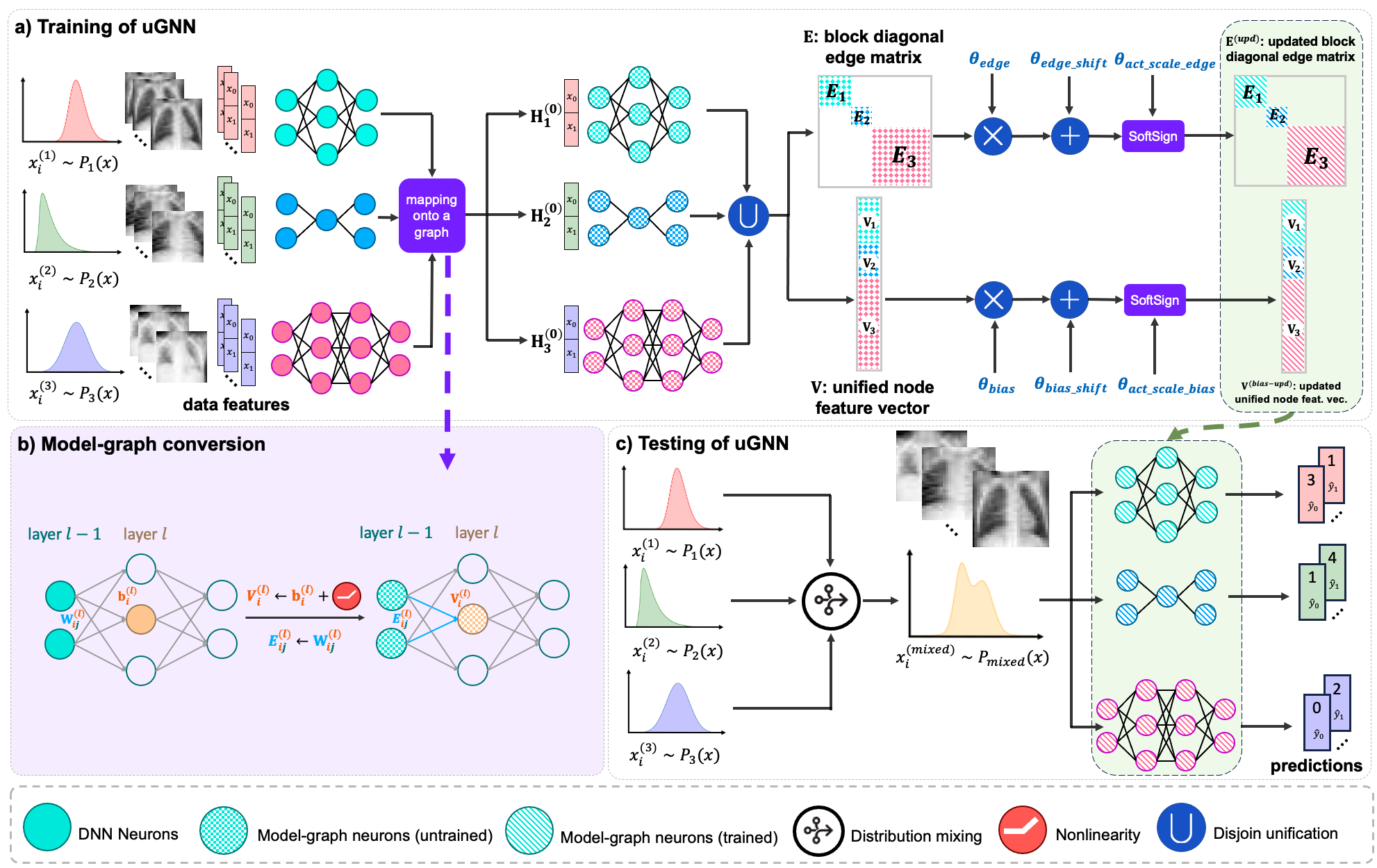}
\caption{Overview of the proposed \texttt{uGNN} framework. \textbf{(a)} \textbf{Training of \texttt{uGNN}:} Input data from multiple distributions ${P_1(x), P_2(x), P_3(x)}$ are fed into heterogeneous DNNs (MLPs and CNNs). Each DNN is mapped onto a model-graph with node features $\mathbf{V}_i$ (biases) and edge features $\mathbf{E}_i$ (weights). A unified graph is constructed through disjoint unification, where shared parameters $(\theta_{\text{edge}}, \theta_{\text{bias}})$ and corresponding shifts rescale and transform the edge and node features. \textbf{(b)} \textbf{Model-graph conversion:} DNN neurons are converted into model-graph neurons with edge features $\mathbf{E}_{ij}^{(l)} \leftarrow \mathbf{W}_{ij}^{(l)}$ and node features $\mathbf{V}_i^{(l)} \leftarrow \mathbf{b}_i^{(l)}$. Nonlinearities are stored in node features. \textbf{(c)} \textbf{Testing of \texttt{uGNN}:} During testing, the models are evaluated on a mixed distribution $P_{mixed}(x)$ to demonstrate improved generalization.}

    \label{fig:main-fig}
\end{figure}

Our \texttt{uGNN} framework consists of three core steps: (i) converting heterogeneous architectures into graph representations, (ii) unifying these graph representations into a common graph-based learning space, and (iii) defining a GNN-driven optimization and training procedure. Steps (ii) and (iii) are illustrated in \textbf{Fig.~\ref{fig:main-fig}-a}, while step (i) is shown in \textbf{Fig.~\ref{fig:main-fig}-b}. Each step is detailed below.

\textbf{Transforming Heterogeneous Architectures to Graphs.} Here, we represent a DNN as a graph with nodes corresponding to DNN neurons and edges corresponding the connections between DNN neurons~\cite{kofinas2024graph}. While node features store the current activation and the bias of the corresponding neuron, edge features store the weight between two neurons. Additionally, we store activation functions as node features, as seen in \textbf{Fig.~\ref{fig:main-fig}-b}, typically chosen from a predefined dictionary of common nonlinearities. In the following section, we first explain the graph transformation of Euclidean models by explaining how we convert MLPs and CNNs into model-graphs. Next, we show how to extend this transformation to non-Euclidean models, i.e., GNNs.

\textbf{MLPs as Graphs.} Consider an MLP with $L$ layers, where each layer $l \in \{1, \dots, L\}$ contains $d_l$ neurons. The weights and biases of the MLP are represented as $\{\mathbf{W}^{(1)}, \mathbf{W}^{(2)}, \dots, \mathbf{W}^{(L)}\}$ and $\{\mathbf{b}^{(1)}, \mathbf{b}^{(2)}, \dots, \mathbf{b}^{(L)}\}$, respectively, where $\mathbf{W}^{(l)} \in \mathbb{R}^{d_l \times d_{l-1}}$ and $\mathbf{b}^{(l)} \in \mathbb{R}^{d_l \times 1}$. The total number of neurons across all layers (including input and output layers) is $n = \sum_{l=0}^{L} d_l$, with $d_0$ represents the number of input neurons. We construct a model-graph $G = (\mathbf{V}, \mathbf{E})$, where $\mathbf{V} \in \mathbb{R}^{n \times d_v}$ node features, and $\mathbf{E} \in \mathbb{R}^{n \times n \times d_e}$ represents edge features. Each node corresponds to a single neuron, and edges represent the weighted connections between neurons in consecutive layers. Importantly, due to the feedforward structure of MLPs, these connections form a \emph{directed acyclic graph (DAG)}, ensuring no circular dependencies exist between nodes.

\noindent{\emph{MLP Node Features.}} As depicted in \textbf{Fig.~\ref{fig:main-fig}-b}, each node corresponds to a neuron in some layer $l$. The feature vector for a node $v^{(l)}_j$, representing the $j$-th neuron in layer $l$, is initialized from its corresponding bias term
$
\mathbf{V}_{i} = \mathbf{b}^{(l)}_{j}
$
where $i = \text{Index}(l, j)$ maps the layer index $l$ and neuron index $j$ within that layer to a unique global node index $i$ in $\mathbf{V}$.

\noindent{\emph{MLP Edge Features.}} For edges, consider the connection from neuron $v^{(l-1)}_{m}$ in layer $(l-1)$ to neuron $v^{(l)}_j$ in layer $l$ as visualized in \textbf{Fig.~\ref{fig:main-fig}-b}. The corresponding weight $\mathbf{W}^{(l)}_{jm}$ serves as the edge feature. Let $i=\text{Index}(l,j)$ be the global node index of neuron $j$ in layer $l$, and $k=\text{Index}(l-1,m)$ the global node index of neuron $m$ in layer $(l-1)$. Then:
$
\mathbf{E}_{ik} = \mathbf{W}^{(l)}_{jm}
$

\noindent{\textbf{CNNs as Graphs.}} We now extend our graph representation from MLPs to CNNs. Consider a CNN architecture consisting of convolutional and fully connected (linear) layers. Each convolutional layer transforms an input feature map of size $C_{\text{in}} \times H_{\text{in}} \times W_{\text{in}}$ into an output feature map $C_{\text{out}} \times H_{\text{out}} \times W_{\text{out}}$, where $C_{\text{in}}$/$C_{\text{out}}$ are the input/output channels and $H_{\text{in}}, W_{\text{in}}, H_{\text{out}}, W_{\text{out}}$ are the input/output spatial dimensions. $L$ denotes the total number of layers (counting both convolutional and fully connected layers), and $n$ the total number of “neuronal units” (i.e., spatial positions across all feature maps at all layers, plus the eventual neurons in fully connected layers).

We construct a model-graph $G = (\mathbf{V}, \mathbf{E})$ for the CNN in a manner analogous to the MLP case. Nodes now correspond to individual spatial locations in each layer’s feature map, and edges represent the connections defined by receptive fields across layers. Formally, $\mathbf{V} \in \mathbb{R}^{n \times d_v}$ denotes the node feature matrix, and $\mathbf{E} \in \mathbb{R}^{n \times n \times d_e}$ denotes the edge feature tensor.

\noindent{\emph{CNN Node Features.}} Each node represents a single neuron at a particular spatial position and channel in a given layer. For an input layer (e.g., the raw image), each pixel position and channel is treated as a node. For a convolutional layer $l$, each output neuron corresponds to a spatial location $(i, j)$ and channel $c_{\text{out}}$, yielding a node $v^{(l)}_{c_{\text{out}}, i, j}$. The bias parameters of the convolutional filter are used to initialize node features. Specifically, if $\mathbf{b}^{(l)} \in \mathbb{R}^{C_{\text{out}}\times1}$ is the bias vector for layer $l$, each node corresponding to output channel $c_{\text{out}}$ inherits its node feature from $\mathbf{b}^{(l)}_{c_{\text{out}}}$. Thus, $\mathbf{V}_{u} = \mathbf{b}^{(l)}_{c_{\text{out}}}, \quad u = \text{Index}(l, c_{\text{out}}, i, j)$
where $\text{Index}(\cdot)$ provides a unique global node index for each spatial location and channel combination. If the network includes fully connected (linear) layers at the end, each neuron in these layers is also treated as a node, whose feature vector is initialized from the corresponding bias term in the linear layer’s bias vector, analogous to the MLP case.

\noindent{\emph{CNN Edge Features.}} Edges capture how each neuron in layer $l$ is computed from a local receptive field of neurons in layer $(l-1)$. For a convolutional layer, the kernel weights $\mathbf{W}^{(l)} \in \mathbb{R}^{C_{\text{out}} \times C_{\text{in}} \times K_{h} \times K_{w}}$ define connections from a $K_{h} \times K_{w}$ patch of neurons in the previous layer’s feature maps to a single output neuron. Thus, for each node $v^{(l)}_{c_{\text{out}}, i, j}$, we link it to nodes representing the patch $(i', j')$ and channel $c_{\text{in}}$ in layer $(l-1)$ that lie within its receptive field. The corresponding edge feature is the kernel weight $\textbf{W}^{(l)}_{c_{\text{out}}, c_{\text{in}}, \Delta i, \Delta j}$, where $(\Delta i, \Delta j)$ indexes the kernel element mapping from position $(i-\Delta i, j-\Delta j)$ in the previous layer to $(i, j)$ in the current layer:
\[
\mathbf{E}_{uv} = \textbf{W}^{(l)}_{c_{\text{out}}, c_{\text{in}}, \Delta i, \Delta j}, \quad u = \text{Index}(l, c_{\text{out}}, i, j), \quad v = \text{Index}(l-1, c_{\text{in}}, i-\Delta i, j-\Delta j)
\]

For subsequent fully connected layers, edges are constructed similarly to the MLP case: each output neuron in the fully connected layer connects to all neurons in the previous layer, and edge features store the corresponding linear weights.

\noindent{\emph{Graph Connectivity and DAG Structure.}} As for MLPs, the CNN’s feedforward nature ensures that the resulting graph is a DAG. Nodes in earlier layers connect forward to nodes in later layers, but no edges exist backward or laterally that would introduce cycles:
\[
\mathbf{A}_{uv} = 
\begin{cases}
1 & \begin{aligned}
   &\text{if $v_u$ is a node in layer $l$, $v_v$ is a node in layer $(l-1)$,}\\
   &\text{and $v_v$ lies within the receptive field of $v_u$,}
   \end{aligned}\\[6pt]
0 & \text{otherwise.}
\end{cases}
\]

\noindent\textbf{GNNs as Graphs.} We now extend our graph representation to GNNs themselves. Unlike MLPs and CNNs, GNNs already operate on graph-structured data, but we can still represent their architectures as meta-graphs for unified learning. In this formulation, each node in the model-graph represents a node in the input graph at a specific layer of the GNN, while edges represent the aggregation of messages across layers.

Consider a GNN with $L$ layers operating on input graphs with $N$ nodes. For each layer $l$, let $\mathbf{H}^{(l)} \in \mathbb{R}^{N \times d_l}$ denote the node embeddings at that layer, where $d_l$ is the embedding dimension. The GNN’s message passing can be abstracted as a sequence of transformations:
$
\mathbf{H}^{(l)} = \sigma \big( \hat{\mathbf{A}} \mathbf{H}^{(l-1)} \mathbf{W}^{(l)} \big)
$
where $\hat{\mathbf{A}}$ is the normalized adjacency matrix (potentially with self-loops), $\mathbf{W}^{(l)}$ is the trainable weight matrix at layer $l$, and $\sigma$ is a nonlinearity such as ReLU.

\emph{GNN Node Features.} In our model-graph, each node represents a tuple $(v, l)$ where $v$ is a node in the input graph and $l$ is a layer index. The node feature combines the bias associated with that layer and the embedding dimension:
$
\mathbf{V}_{(v,l)} = \mathbf{b}^{(l)}
$
where $\mathbf{b}^{(l)}$ is the bias vector of layer $l$ (broadcasted across all input nodes).

\emph{GNN Edge Features.} Edges in the model-graph represent message passing between nodes across layers. For nodes $u$ and $v$ in the input graph connected by an edge, we add edges in the model-graph between $(u,l-1)$ and $(v,l)$ for each layer $l$. We define the edge features correspond to the weight matrix $\mathbf{W}^{(l)}$ applied during aggregation as 
$
\mathbf{E}_{(v,l),(u,l-1)} = \mathbf{W}^{(l)}
$

\noindent\textbf{Unifying Model-Graphs into a Common Learning Space.}
Consider a collection of $m$ heterogeneous models $\{M_1, M_2, \dots, M_m\}$, each represented by its own model-graph $(\mathbf{V}_i, \mathbf{E}_i)$ as described above. We construct a unified representation by forming a disjoint union of all these graphs (\textbf{Fig.~\ref{fig:main-fig}-a}):
$
\mathbf{V} = \bigcup_{i=1}^m \mathbf{V}_i, \quad 
\mathbf{E} = \bigcup_{i=1}^m \mathbf{E}_i.
$
This operation creates a larger graph $G=(\mathbf{V}, \mathbf{E})$ containing all nodes and edges from each individual model-graph, without introducing edges between different models. Consequently, $G$ can be viewed as a block-diagonal adjacency structure, where each block corresponds to the topology of a single model. By training a GNN on this unified graph $G$, we process all subgraphs (each corresponding to a different model) simultaneously with a single set of shared parameters. This enables the GNN to learn a common representation space, potentially capturing model-invariant properties and facilitating knowledge transfer among the diverse set of model architectures.

\noindent\textbf{Parametrization.}  
Each model’s edges correspond to parameterized connections (weights) and each node corresponds to a neuron with an associated bias. To replicate these computations, we introduce shared learnable parameters $(\theta_{\text{edge}}, \theta_{\text{bias}}, \theta_{\text{edge_shift}}, \theta_{\text{bias_shift}},\theta_{\text{act_scale_edge}}, \theta_{\text{act_scale_bias}})$ that adjust and transform the original edges and biases from all models simultaneously. The pipeline of these parameters are visually depicted in \textbf{Fig.~\ref{fig:main-fig}-a}. Specifically, let $\theta_{\text{edge}} \in \mathbb{R}^{k_{\text{edge}}}$ and $\theta_{\text{edge_shift}} \in \mathbb{R}^{k_{\text{edge}}}$ be learnable parameters used to rescale and shift the edge features (corresponding to underlying model weights). We randomly partition edges into $k_{\text{edge}}$ groups, and each edge group shares a single scalar scale and shift parameter. For an edge $(u,v)$ belonging to group $g(u,v)$, we define the updated edge features as
    $
    \mathbf{E}_{u,v}^{(\text{upd})} = \operatorname{SoftSign}\bigl(\mathbf{E}_{u,v} \cdot \theta_{\text{edge}}^{(g(u,v))} + \theta_{\text{edge_shift}}^{(g(u,v))},\, \theta_{\text{act_scale_edge}}\bigr)
    $
    where $\operatorname{SoftSign}(x,s)$ is an element-wise nonlinearity scaled by $s$ defined as $SoftSign(x,s) = \frac{s \cdot x}{s + |x|}$. Note that $\mathbf{E}_{u,v}$ denotes the original edge feature (e.g., a weight) and $\theta_{\text{act_scale_edge}}$ is a scalar controlling the $\operatorname{softsign}$ scaling. Similarly, node biases are grouped into $k_{\text{bias}}$ groups. Each node bias group has its own scale and shift parameters $(\theta_{\text{bias}}, \theta_{\text{bias_shift}}) \in \mathbb{R}^{k_{\text{bias}}}$. For a node $v$ in group $h(v)$, we update its bias as
    $
    \mathbf{V}_{v}^{(\text{bias-upd})} = \operatorname{SoftSign}\bigl(\mathbf{V}_{v} \cdot \theta_{\text{bias}}^{(h(v))} + \theta_{\text{bias_shift}}^{(h(v))},\, \theta_{\text{act_scale_bias}}\bigr)
    $
    where $\mathbf{V}_{v}$ is the original bias feature of the node $v$ (derived from the underlying model parameters), and $\theta_{\text{act_scale_bias}}$ adjusts the scaling of the $\operatorname{SoftSign}$ function. By applying these transformations, the GNN aligns all underlying model parameters (edges and biases) into a common parametrization that can be adjusted during training.

\noindent\textbf{Message Passing and Activation.}  
After updating edges and node biases as described, the forward pass emulates the feedforward computation of the underlying models layer-by-layer. Let $\mathbf{H}^{(\ell)} \in \mathbb{R}^{n \times 1}$ represent the node activations in layer $\ell$, where layers are implicitly defined by the topological ordering in $G$. As illustrated in \textbf{Fig.~\ref{fig:main-fig}-a,} for the first layer, we set
$
\mathbf{H}^{(0)} = \mathbf{X}
$ where $\mathbf{X}$ are input features for input nodes (and zeros for other nodes until their layer’s turn). At each subsequent layer $\ell > 0$, we use the updated edge features to aggregate incoming signals $\mathbf{Z}^{(\ell)}_v = \sum_{v:\mathbf{A}_uv = 1} \mathbf{E}_{uv}^{\text{(upd)}} \cdot \mathbf{H}^{(\ell-1)}_u $ where $\mathbf{E}_{uv}^{\text{(upd)}}$ are the updated edge features from node $u$ to $v$, and $\mathbf{H}^{(\ell-1)}_u$ is the activation of node $u$ from the previous layer. This operation is analogous to performing a matrix multiplication of weights with activations in an MLP or applying convolution-like operations in a CNN. We then add the updated node bias $
\mathbf{Z}^{(\ell)}_v \leftarrow \mathbf{Z}^{(\ell)}_v + \mathbf{V}_{v}^{\text{(bias-upd)}}
$ where $\mathbf{V}_v^{\text{(bias-upd)}}$ is the updated bias for node $i$.

\noindent\textbf{Layer-by-Layer Computation and Model Emulation.}  
By iterating this process from $\ell = 1$ up to the deepest layer present in any model subgraph, we replicate the forward pass of each underlying model within this unified GNN framework. The parameterized transformations $(\theta_{\text{edge}}, \theta_{\text{bias}})$ and associated shifts ensure that our \texttt{uGNN} can adaptively rescale and align parameters so that the resulting node outputs match the computations of the target models.

\noindent\textbf{Loss Computation and Training.}  
Following the \texttt{uGNN} forward pass and the computation of the node activations at the final layer, $\mathbf{H}^{(L_i)}$, we can extract the predictions of each original model $M_i$. Since all models share the same classification task, final-layer activations , $\mathbf{H}^{(L_i)}$, corresponds to the predicted logits by model $M_i$. Let $\mathbf{Y}^{(i)}$ be the ground-truth labels associated with model $M_i$’s data. For classification, we use a standard loss function such as cross-entropy:
$
\mathcal{L}_i = -\sum_{c} Y^{(i)}_c \log \hat{Y}^{(i)}_c,
$ where $\hat{\mathbf{Y}}^{(i)}$ are the softmax-normalized predictions derived from $\mathbf{H}^{(L_i)}$. Each $\mathcal{L}_i$ measures how well the \texttt{uGNN}’s predictions match the ground-truth labels for the $i$-th model’s subgraph. To train the GNN so that it can accurately emulate all $m$ original models' computations, we combine the individual losses into a single scalar objective by computing a weighted sum $
\mathcal{L} = \sum_{i=1}^m \alpha_i \mathcal{L}_i, \quad \text{where} \quad \sum_{i=1}^m \alpha_i = 1 \text{ and } \alpha_i \geq 0$. Here, $\alpha_i$ represents the weight assigned to the $i$-th model's loss, reflecting its relative importance during training. These weights can be learned or set manually based on specific criteria, such as dataset size or task complexity. We minimize $\mathcal{L}$ with respect to all GNN parameters $(\theta_{\text{edge}}, \theta_{\text{bias}}, \theta_{\text{edge_shift}}, \theta_{\text{bias_shift}},\theta_{\text{act_scale_edge}}, \theta_{\text{act_scale_bias}})$  using standard gradient-based optimization. By doing so, our \texttt{uGNN} learns a unified parametrization that can accurately reconstruct the forward pass of each original model, leading to a single learned representation that successfully performs the classification task for all model-graphs.

\section{Experiments \& Results}

To evaluate the generalization capabilities of our unified learning paradigm, we design our experimental setup as follows. 
During training, each model is associated with a particular distribution:
$
x_i^{(k)} \sim P_k(x),
$
where $P_k(x)$ denotes the $k$-th distributional cluster derived from the data. At testing time, as illustrated in \textbf{Fig.~\ref{fig:main-fig}-c}, each model encounters test samples drawn from a mixture of distributions:
$
x_i^{(\text{mixed})} \sim P_{\text{mixed}}(x), \quad \text{where} \; P_{\text{mixed}}(x) = \sum_{k=1}^{K} P_k(x),
$ and $K$ is the number of clusters. This setup directly tests the ability of models to generalize beyond their individually seen distributions. 

We consider two training scenarios: \textbf{(1) Individual Training:} Each model is trained independently on a single cluster, without knowledge of the other clusters (i.e., distributions). \textbf{(2) \texttt{uGNN} Training:} Using our unified GNN framework, we \emph{jointly} train multiple model-graphs, each independently accessing data with a distinct distribution (\textbf{Fig.\ref{fig:main-fig}-a}). Although each model-graph individually accesses only one specific distribution, the shared parameters $(\theta_{\text{edge}}, \theta_{\text{bias}})$ within the \texttt{uGNN} facilitate indirect knowledge transfer across distributions. In this manner, the \texttt{uGNN} leverages signals from multiple distributions without explicitly sharing them between models, potentially enhancing generalization.

\begin{figure}[ht]
    \centering
    \includegraphics[width=0.75\linewidth]{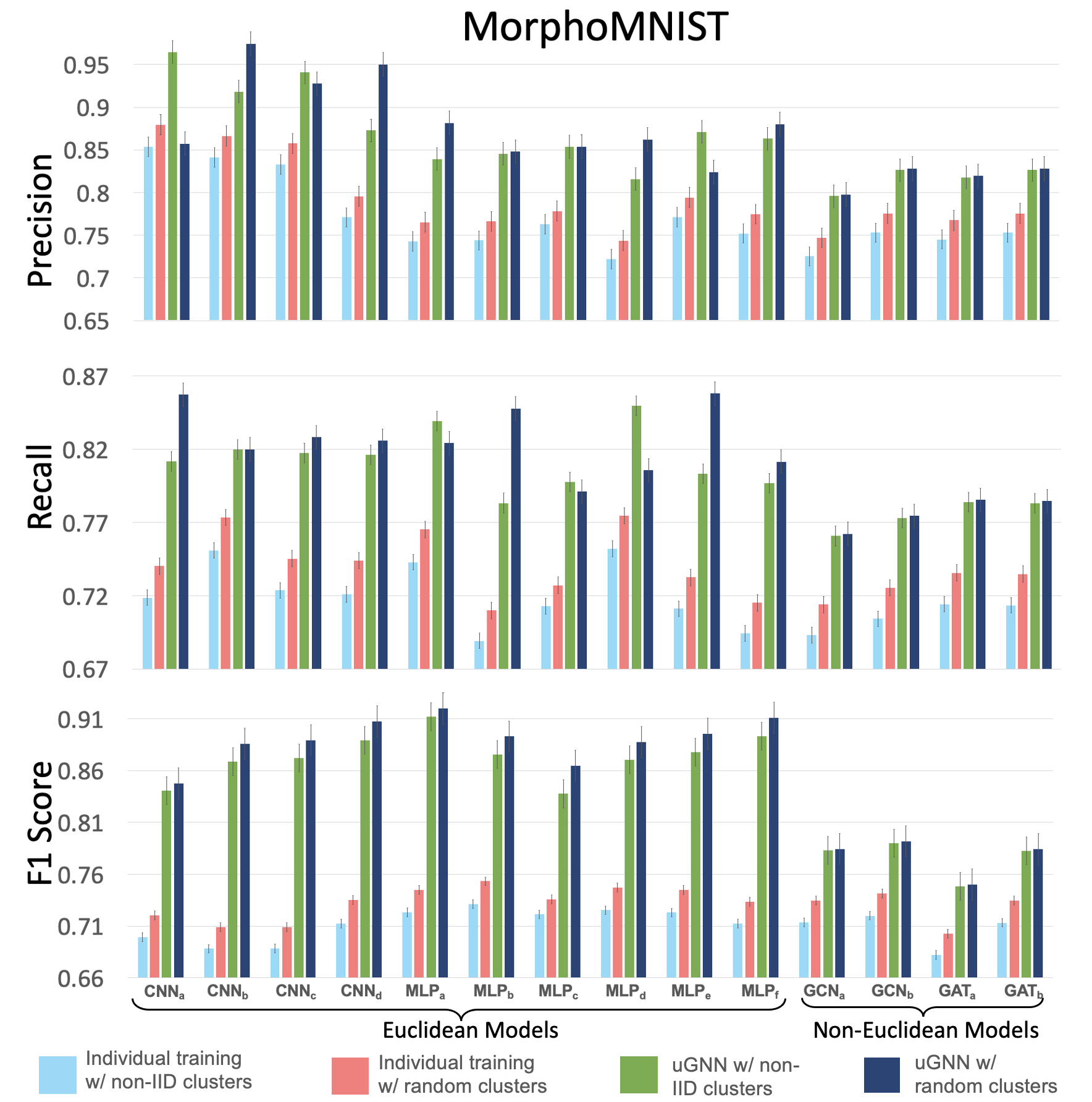}
    \caption{Comparison of testing performance on the MorphoMNIST dataset between the baseline approach and our \texttt{uGNN} across six Euclidean and four non-Euclidean models. Each architecture has been trained on either plain, thinned or thickened perturbations.}
    \label{fig:morpho_results}
\end{figure}

\begin{figure}[ht]
    \centering
    \includegraphics[width=\linewidth]{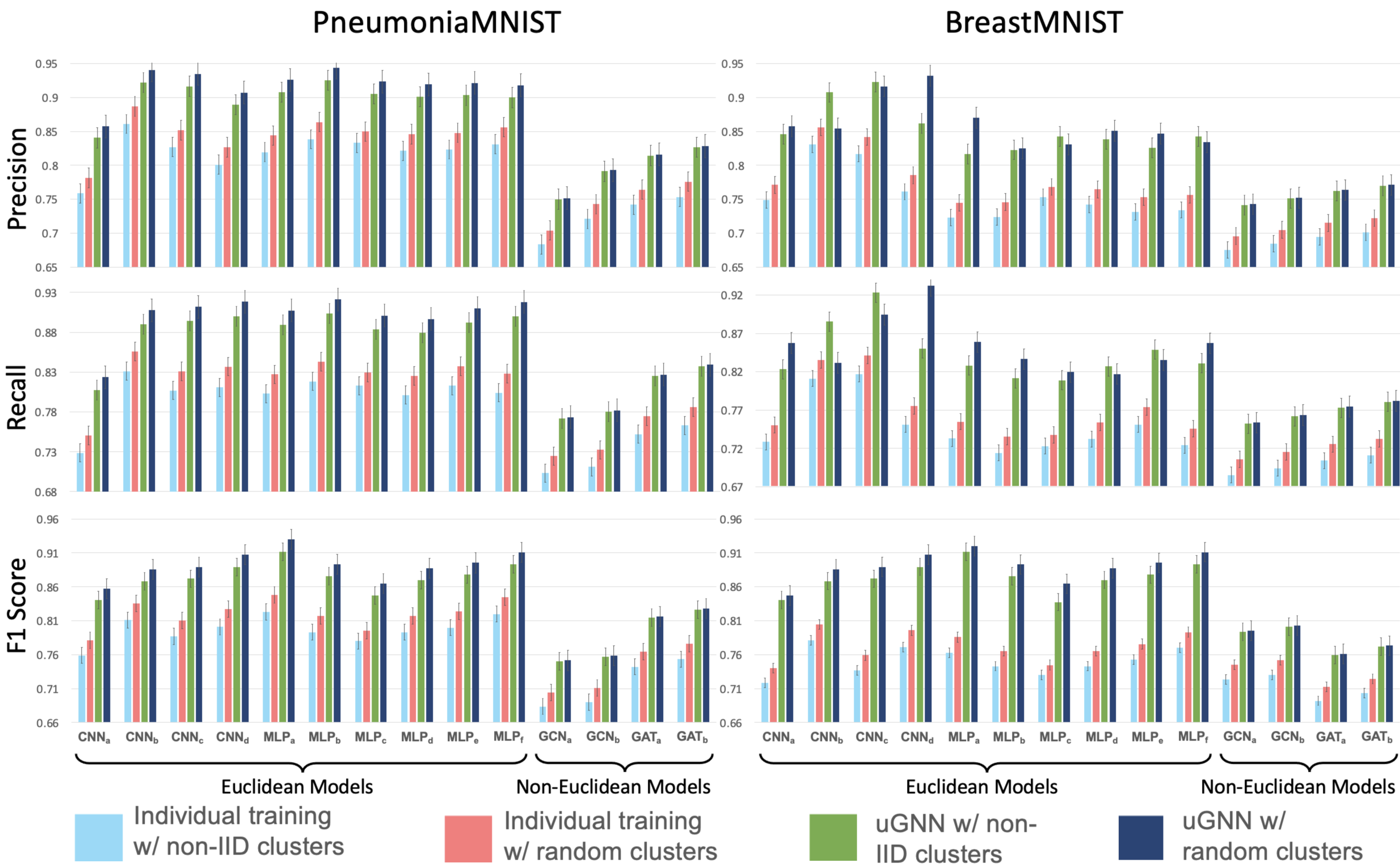}
\caption{Comparison of testing performance for six Euclidean and four non-Euclidean models on the PneumoniaMNIST and BreastMNIST datasets, evaluated using the baseline approach and our \texttt{uGNN}.}
    \label{fig:med_mnist_results}
\end{figure}

Our experiments focus on the BreastMNIST and PneumoniaMNIST datasets from~\textbf{MedMNIST}~\cite{yang2023medmnist} and \textbf{MorphoMNIST}~\cite{castro2019morphomnist}. For each dataset, we have two clustering setup: \textbf{(1) Non-IID}: To generate non-IID clusters, we first reduce the dimensionality of the $28 \times 28$ images using Principal Component Analysis (PCA) with 50 components, followed by $k$-means clustering on the PCA-transformed features with a distinct \emph{random seed} for each fold, ensuring diversity and reproducibility in the clustering process. The resulting clusters represent distinct distributions. \textbf{(2) Random}: In random clustering, we randomly create clusters from samples in the dataset so that clusters do not exhibit any engineered distribution shift between them.

During training, each cluster is assigned to a separate model-graph, allowing the \texttt{uGNN} framework to unify and refine parameters across these diverse data subsets. 

In all our experiments, we set $k_{\text{edge}}$ and $k_{\text{bias}}$ to the total number of edges and nodes in the unified graph $G$, respectively. As a result, each edge and bias group is associated with a single parameter. To ensure a fair comparison, we use the same training setup for both individual models and the \texttt{uGNN} framework: the AdamW optimizer with identical learning rate and weight decay settings, the same number of training epochs, and the same learning rate scheduler. The initialization of underlying models is also identical for both approaches. To ensure stable training, we initialize both the underlying models and the \texttt{uGNN} parameters carefully. We initialize an underlying model identically for both \texttt{uGNN} and baseline training versions. For the \texttt{uGNN} parameters, aggressive schemes like Gaussian or Kaiming initialization~\cite{he2015delving} destabilize training by causing excessive updates to edge and node features. Therefore, we adopt the following conservative initialization: $
\theta_{\text{edge}} = \mathbf{1}_{k_{\text{edge}}}, \quad 
\theta_{\text{edge\_shift}} = \mathbf{0}_{k_{\text{edge}}}, \quad 
\theta_{\text{bias}} = \mathbf{1}_{k_{\text{bias}}}, \quad 
\theta_{\text{bias\_shift}} = \mathbf{0}_{k_{\text{bias}}}, \quad 
\theta_{\text{edge\_act\_scale}} = 1, \quad 
\theta_{\text{bias\_act\_scale}} = 1
$. This initialization ensures minimal initial updates to edge and node features.

As shown in \textbf{Fig.~\ref{fig:morpho_results}}, we evaluate on \textbf{MorphoMNIST}~\cite{castro2019morphomnist}, where DNNs are trained on plain, thinned, or thickened digits, and tested on a mixed distribution. \texttt{uGNN} consistently improves generalization across all models compared to individual training. \textbf{Fig.~\ref{fig:med_mnist_results}} compares individual training with \texttt{uGNN} on \textbf{PneumoniaMNIST} and \textbf{BreastMNIST}. \texttt{uGNN} outperforms individual training in all cases, demonstrating significant performance gains on real-world clinical datasets—highlighting its potential utility in medical imaging applications. Furthermore, we observe that models trained on random client clusters perform better than those trained on non-IID clusters. This is because random clustering does not introduce additional domain shifts beyond the dataset's inherent distribution, whereas non-IID clustering exacerbates domain discrepancies by design.

\section{Discussion \& Conclusion}
In this work, we introduced a unified learning framework that brings together heterogeneous neural network architectures—namely MLPs, CNNs, and GNNs—within a shared graph-based representation for joint training under a common parameter space. By converting diverse architectures into graph structures, our approach enables indirect knowledge transfer across models, even when they operate in fundamentally different (Euclidean vs. non-Euclidean) spaces. This leads to improved generalization, particularly under domain shifts and distributional heterogeneity. Our framework proves effective in both homogeneous and heterogeneous model settings by harmonizing architectural differences and enabling collaborative learning. Looking ahead, we aim to extend this framework to support \emph{multi-modality} and \emph{multi-task learning}, paving the way for unified, distribution- and architecture-agnostic learning paradigms. Such capabilities are especially promising for collaborative settings like federated learning, with high-impact applications in domains such as precision medicine.

\bibliographystyle{splncs}
\bibliography{references}

\end{document}